\documentclass[cameraready]{Interspeech}
\usepackage[T1]{fontenc}
\usepackage[utf8]{inputenc}
\usepackage[dvipsnames,table]{xcolor}
\usepackage[noabbrev]{cleveref}
\usepackage{silence,caption,subcaption,xspace,multirow,microtype,threeparttable}
\newcommand{\dev}{dev\xspace}
\newcommand{\test}{test\xspace}
\newcommand{\benchmark}{DiscoPhon\xspace}
\newcommand{\sil}{\texttt{SIL}\xspace}
\newcommand{\prob}{\mathbb{P}\xspace}
\renewcommand{\paragraph}[1]{{\noindent\small$\blacktriangleright$} \textbf{#1}}

\title{\benchmark: Benchmarking the Unsupervised Discovery of \texorpdfstring{\\}{} Phoneme Inventories With Discrete Speech Units}
\author[affiliation={1},orcid=0000-0002-9377-9150]{Maxime}{Poli}
\author[affiliation={1},orcid=0009-0004-7077-4332]{Manel}{Khentout}
\author[affiliation={1},orcid=0009-0002-0450-8381]{Angelo}{Ortiz Tandazo}
\author[affiliation={2},orcid=0000-0001-9603-953X]{Ewan}{Dunbar}
\author[affiliation={1},orcid=0000-0002-8423-5880]{Emmanuel}{Chemla}
\author[affiliation={1},orcid=0000-0002-7814-2952]{Emmanuel}{Dupoux}
\address{$^1$ LSCP, ENS, EHESS, CNRS, PSL University, France \\ $^2$ University of Toronto, Canada}
\email{benchmarks@cognitive-ml.fr}
\keywords{self-supervised learning, discrete speech units, multilinguality}

\begin{document}
\maketitle

\begin{abstract}
We introduce \benchmark, a multilingual benchmark for evaluating unsupervised phoneme discovery from discrete speech units. \benchmark covers 6 \dev and 6 \test languages, chosen to span a wide range of phonemic contrasts. Given only 10 hours of speech in a previously unseen language, systems must produce discrete units that are mapped to a predefined phoneme inventory, through either a many-to-one or a one-to-one assignment. The resulting sequences are evaluated for unit quality, recognition and segmentation. We provide four pretrained multilingual HuBERT and SpidR baselines, and show that phonemic information is available enough in current models for derived units to correlate well with phonemes, though with variations across languages.
\end{abstract}

\section{Introduction and related work}
Imagine a linguist arriving in a new community to document the local language. One of the first tasks is to establish its phoneme inventory (consonants and vowels). Could an automatic system help in this process? Half of the world's languages are at risk of disappearing by the end of the century \cite{moseley2010atlas}, many before they can be documented. Several studies have addressed automatic phoneme inventory discovery \cite{kempton2014discovering,wang-etal-2022-self,7953148}. Here, we propose a benchmark that leverages self-supervised speech representation learning (SSL) to accelerate research in this area.

Discovering the correct set of phonemes for a language is crucial: encoding the wrong categories distorts or erases contrasts between words. A natural approach is to learn a mapping from speech to a finite set of categories directly from raw audio. This problem can be cast in terms of automatic discovery of discrete speech units, a line of research related to spoken term discovery \cite{park2008unsupervised,ludusan-etal-2014-bridging,kamper2016unsupervised,kamper2022word}. Recent work has shown that self-supervised models encode phonetic and phonemic information at readily accessible levels in their representations \cite{deseyssel22_interspeech,wells2022phonetic,choi24b_interspeech}. Combining such models with a quantization step, like $K$-means vector quantization, should therefore yield units that retain this information \cite{abdullah2023informationtheoretic,yeh2024estimating}. These resulting pseudo-linguistic units can then serve as input to language models \cite{lakhotia-etal-2021-generative,nguyen2020zeroresourcespeechbenchmark,poli2024improving}, enabling semantic modeling of speech disentangled from speaker and acoustic characteristics \cite{borsos2023audiolm,nguyen-etal-2025-spirit}. For this to work, however, the units must abstract away from low-level acoustic properties and capture phoneme-level information, which is typically not the case for approaches such as neural codecs \cite{nguyen23_interspeech}. Whether such units generalize to typologically diverse languages remains largely untested. Our benchmark is designed to evaluate precisely this capability, in a way that integrates naturally with current SSL-based tokenization pipelines.
 
Past benchmarks \cite{dunbar2017zeroa,dunbar2021zero,dunbar2022selfsupervised} have primarily focused on the evaluation of SSL representations that carry linguistic contrast as measured through ABX discrimination methods. Here, we go one step further and focus on deriving phoneme-like discrete units from these continuous representations from limited data (\SI{10}{\hour}), without supervision or annotations. This can be framed as an unsupervised clustering problem in which the discovered units should ideally correspond to the phoneme categories a linguist would use to describe the target language. This immediately raises the issue of how many phonemes a language contains, which is difficult to answer, even for linguists, due to a long tail of allophones and quasi-phonemes \cite{clements1985geometry}.

\begin{figure}
    \centering\includegraphics{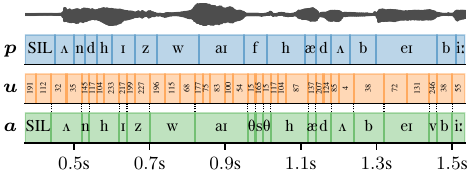}
     \caption{\textbf{Overview of the streams in the evaluation pipeline.} The model maps a waveform to units $\bm{u}$, evaluated against the gold phones $\bm{p}$ via $\textnormal{PNMI}(\bm{p}, \bm{u})$. An assignment, here many-to-one, then assigns units to phones $\bm{a}$, assessed for recognition with $\textnormal{PER}(\bm{p}, \bm{a})$ and segmentation with $R\textnormal{-value}(\bm{p}, \bm{a})$ and $F_1(\bm{p}, \bm{a})$.\label{fig:streams}}
\end{figure}

\begin{table*}
    \addtolength{\tabcolsep}{-0.25em}
    \caption{\textbf{Phonemic categories considered in the benchmark.} \dev languages: German (41 phonemes), Swahili (29), Tamil (29), Thai (40), Turkish (27), and Ukrainian (35). \test languages: Basque (29 phonemes), English (39), French (34), Japanese (42), Mandarin Chinese (42), and Wolof (39). German, English, French, and Wolof data from \cite{dunbar2017zeroa,gauthier2016collectinga}, other languages from \cite{ardila2020commona,ahn2022voxcommunis}.\label{tab:phonemes}} 
   \centering\begingroup
\eightpt
\newcommand{\dr}[1]{\multirow{2}{*}{#1}}
\setlength{\defaultaddspace}{0.2em}

\begin{tabular}{lllllllll}
\toprule
    & Fricative & Affricate & Plosive & Vibrant & Nasal & Approximant & Monophthong & Diphthong \\
\midrule
\multicolumn{9}{c}{\cellcolor{gray!10} \dev languages}  \\
\dr{\textbf{German}} & \textipa{f}, \textipa{s}, \textipa{S}, \textipa{\c{c}}, \textipa{x}, \textipa{h} & \dr{\textipa{\t{ts}}} & \dr{\textipa{p}, \textipa{t}, \textipa{k}, \textipa{b}, \textipa{d}, \textipa{\textscriptg}} & \dr{\textipa{r}, \textipa{\textfishhookr}} & \dr{\textipa{m}, \textipa{n}, \textipa{\ng}} & \dr{\textipa{l}, \textipa{j}} &\textipa{I}, \textipa{i:}, \textipa{Y}, \textipa{y:}, \textipa{U}, \textipa{u:}, \textipa{e:}, \textipa{\o:} & \dr{\textipa{a\textsubarch{I}}, \textipa{a\textsubarch{U}}} \\
&  \textipa{v}, \textipa{z} & & & & & & \textipa{@}, \textipa{o:}, \textipa{E}, \textipa{E:}, \textipa{\textrevepsilon}, \textipa{\oe}, \textipa{O}, \textipa{a}, \textipa{a:} \\
\addlinespace

\dr{\textbf{Swahili}} & \textipa{f}, \textipa{T}, \textipa{s}, \textipa{S}, \textipa{h} & \dr{\textipa{\t{tS}}} & \dr{\textipa{p}, \textipa{t}, \textipa{k}, \textipa{b}, \textipa{d}, \textipa{\textbardotlessj}, \textipa{\textscriptg}} & \dr{\textipa{\textfishhookr}} & \dr{\textipa{m}, \textipa{n}, \textipa{\textltailn}} & \dr{\textipa{w}, \textipa{l}, \textipa{j}} & \dr{\textipa{i}, \textipa{u}, \textipa{E}, \textipa{O}, \textipa{A}} \\
& \textipa{v}, \textipa{D}, \textipa{z}, \textipa{G} \\
\addlinespace

\textbf{Tamil} & \textipa{s}, \textipa{\textrtails} & \textipa{\t{tS}}, \textipa{\t{dZ}} & \textipa{p}, \textipa{\textsubbridge{t}}, \textipa{\textrtailt}, \textipa{k} & \textipa{\textfishhookr} & \textipa{m}, \textipa{n}, \textipa{\textrtailn}, \textipa{\textltailn}, \textipa{\ng} & \textipa{V}, \textipa{l}, \textipa{\textrtaill}, \textipa{\textturnrrtail}, \textipa{j} &\textipa{i}, \textipa{i:}, \textipa{u}, \textipa{u:}, \textipa{e}, \textipa{e:}, \textipa{o}, \textipa{o:}, \textipa{a}, \textipa{a:} \\
\addlinespace

\dr{\textbf{Thai}} & \dr{\textipa{f}, \textipa{s}, \textipa{h}} & \dr{\textipa{\t{tC}}, \textipa{\t{tC}\super{h}}} & \textipa{p}, \textipa{t}, \textipa{k}, \textipa{P}, \textipa{b}, \textipa{d} & \dr{\textipa{r}} & \dr{\textipa{m}, \textipa{n}, \textipa{\ng}} & \dr{\textipa{w}, \textipa{l}, \textipa{j}} & \multirow{2}{10em}{\textipa{i}, \textipa{i:}, \textipa{W}, \textipa{W:}, \textipa{u}, \textipa{u:}, \textipa{e:}, \textipa{\textramshorns}, \textipa{\textramshorns:} \textipa{o}, \textipa{o:}, \textipa{E}, \textipa{E:}, \textipa{O:}, \textipa{a}, \textipa{a:}} & \dr{\textipa{i:\textsubarch{a}}, \textipa{W:\textsubarch{a}}, \textipa{u:\textsubarch{a}}} \\
& & & \textipa{p\super{h}}, \textipa{t\super{h}}, \textipa{k\super{h}} \\
\addlinespace

\textbf{Turkish} & \textipa{f}, \textipa{s}, \textipa{S}, \textipa{h},\textipa{v}, \textipa{z} & \textipa{\t{tS}}, \textipa{\t{dZ}} & \textipa{p}, \textipa{t}, \textipa{k}, \textipa{b}, \textipa{d}, \textipa{\textscriptg} & \textipa{\textfishhookr} & \textipa{m}, \textipa{n} & \textipa{l}, \textipa{j} & \textipa{i}, \textipa{y}, \textipa{W}, \textipa{u}, \textipa{e}, \textipa{o}, \textipa{\oe}, \textipa{a} \\
\addlinespace

\dr{\textbf{Ukrainian}} & \textipa{f}, \textipa{s}, \textipa{s\super{j}}, \textipa{S} & \dr{\textipa{\t{ts}}, \textipa{\t{ts}\super{j}}, \textipa{\t{tS}}, \textipa{\t{dZ}}} & \dr{\textipa{p}, \textipa{t}, \textipa{t\super{j}}, \textipa{k}, \textipa{b}, \textipa{d}, \textipa{d\super{j}}} & \dr{\textipa{r}, \textipa{r\super{j}}} & \dr{\textipa{m}, \textipa{n}, \textipa{n\super{j}}} & \dr{\textipa{V}, \textipa{l}, \textipa{l\super{j}}, \textipa{j}} & \dr{\textipa{i}, \textipa{I}, \textipa{u}, \textipa{E}, \textipa{O}, \textipa{a}} \\
& \textipa{x}, \textipa{z}, \textipa{z\super{j}}, \textipa{Z}, \textipa{H}\\

\midrule
\multicolumn{9}{c}{\cellcolor{gray!10} \test languages}  \\
\textbf{Basque} & \textipa{f}, \textipa{\textsubsquare{s}}, \textipa{\textinvsubbridge{s}}, \textipa{S} & \textipa{\t{t\textsubsquare{s}}}, \textipa{\t{t\textinvsubbridge{s}}}, \textipa{\t{tS}} & \textipa{p}, \textipa{t}, \textipa{k}, \textipa{b}, \textipa{d}, \textipa{\textscriptg} & \textipa{r}, \textipa{\textfishhookr} & \textipa{m}, \textipa{n} & \textipa{l}, \textipa{j} & \textipa{i}, \textipa{u}, \textipa{e}, \textipa{o}, \textipa{a} & \textipa{a\textsubarch{i}}, \textipa{a\textsubarch{u}}, \textipa{e\textsubarch{i}}, \textipa{e\textsubarch{u}}, \textipa{o\textsubarch{i}} \\
\addlinespace

\dr{\textbf{English}} & \textipa{f}, \textipa{T}, \textipa{s}, \textipa{S}, \textipa{h} & \dr{\textipa{\t{tS}}, \textipa{\t{dZ}}} & \dr{\textipa{p}, \textipa{t}, \textipa{k}, \textipa{b}, \textipa{d}, \textipa{\textscriptg}} & & \dr{\textipa{m}, \textipa{n}, \textipa{\ng}} & \dr{\textipa{w}, \textipa{l}, \textipa{\textturnr}, \textipa{j}} & \dr{\textipa{I}, \textipa{i:}, \textipa{U}, \textipa{u:}, \textipa{O:}, \textipa{E}, \textipa{\textrhookrevepsilon}, \textipa{2}, \textipa{\ae}, \textipa{A:}} & \dr{\textipa{aI}, \textipa{aU}, \textipa{eI}, \textipa{oU}, \textipa{OI}} \\
& \textipa{v}, \textipa{D}, \textipa{z}, \textipa{Z} & &  & & & &   \\
\addlinespace

\dr{\textbf{French}} & \textipa{f}, \textipa{s}, \textipa{S} & & \dr{\textipa{p}, \textipa{t}, \textipa{k}, \textipa{b}, \textipa{d}, \textipa{\textscriptg}} & & \dr{\textipa{m}, \textipa{n}, \textipa{\textltailn}} & \dr{\textipa{w}, \textipa{l}, \textipa{j}} & \textipa{i}, \textipa{y}, \textipa{u}, \textipa{e}, \textipa{\o}, \textipa{@}, \textipa{o}, \textipa{E}, \textipa{\oe}, \textipa{O}, \textipa{a} & \\
& \textipa{v}, \textipa{z}, \textipa{Z}, \textipa{K} & & & & & & \textipa{\~E}, \textipa{\~\oe}, \textipa{\~O}, \textipa{\~A} \\
\addlinespace

\dr{\textbf{Japanese}} & \textipa{F}, \textipa{s}, \textipa{s:}, \textipa{C}, \textipa{C:} & \dr{\textipa{\t{ts}}, \textipa{\t{dz}}, \textipa{\t{tC}}, \textipa{\t{d\textctz}}} & \textipa{p},\textipa{t}, \textipa{k}, \textipa{b}, \textipa{d}, \textipa{\textscriptg}  & \dr{\textipa{\textfishhookr}} & \textipa{m}, \textipa{n}, \textipa{\textltailn}, \textipa{\ng}, \textipa{\;N} & \dr{\textipa{w}, \textipa{j}} & \dr{\textipa{i}, \textipa{i:}, \textipa{W}, \textipa{W:},  \textipa{e}, \textipa{e:}, \textipa{o}, \textipa{o:}, \textipa{a}, \textipa{a:}} \\
& \textipa{\c{c}}, \textipa{h}, \textipa{z}, \textipa{\textctz} & & \textipa{p:}, \textipa{t:}, \textipa{k:} & & \textipa{m:}, \textipa{n:} \\
\addlinespace

\textbf{Mandarin} & \dr{\textipa{f}, \textipa{s}, \textipa{\textrtails}, \textipa{\textctc}, \textipa{x}} & \textipa{\t{ts}}, \textipa{\t{\textrtailt\textrtails}}, \textipa{\t{t\textctc}} & \dr{\textipa{p}, \textipa{p\super{h}}, \textipa{t}, \textipa{t\super{h}}, \textipa{k}, \textipa{k\super{h}}} & & \dr{\textipa{m}, \textipa{n}, \textipa{\ng}} & \dr{\textipa{w}, \textipa{\textturnh}, \textipa{l}, \textipa{j}, \textipa{\s{\textturnr}}, \textipa{\textturnrrtail}, \textipa{\s{\textturnrrtail}}} & \dr{\textipa{i}, \textipa{y}, \textipa{U}, \textipa{u}, \textipa{e}, \textipa{\textramshorns}, \textipa{o}, \textipa{@}, \textipa{E}, \textipa{a}, \textipa{A}} & \dr{\textipa{a\textsubarch{i}}, \textipa{a\textsubarch{u}}, \textipa{e\textsubarch{i}}, \textipa{o\textsubarch{u}}} \\
\textbf{Chinese} & & \textipa{\t{ts}\super{h}},\textipa{\t{\textrtailt\textrtails}\super{h}}, \textipa{\t{t\textctc}\super{h}} \\
\addlinespace

\dr{\textbf{Wolof}} & \dr{\textipa{f}, \textipa{s}, \textipa{x}} & & \textipa{p}, \textipa{t}, \textipa{c}, \textipa{k}, \textipa{q}, \textipa{b}, \textipa{\super{m}b} & \dr{\textipa{r}} & \dr{\textipa{m}, \textipa{n}, \textipa{\textltailn}, \textipa{\ng}} & \dr{\textipa{w}, \textipa{l}, \textipa{j}} & \textipa{i}, \textipa{u}, \textipa{e}, \textipa{o}, \textipa{@}, \textipa{E}, \textipa{O}, \textipa{a}  \\
& & & \textipa{d}, \textipa{\super{n}d}, \textipa{\textbardotlessj}, \textipa{\super{\textltailn}\textbardotlessj}, \textipa{\textscriptg}, \textipa{\super{\ng}\textscriptg} & & & & \textipa{i:}, \textipa{u:}, \textipa{e:}, \textipa{o:}, \textipa{E:},  \textipa{O:}, \textipa{a:} \\
\bottomrule
\end{tabular}
\endgroup
\end{table*}

This is why we structured this challenge into two tracks. In the first track, we allow participants to discover a large number of units (arbitrarily fixed at 256), and the evaluation assumes a \textit{many-to-one} mapping between these units (essentially allophonic variants) and the ground truth phonemes. This allows participants to focus on the phonetic purity of the units, without having to worry about how to group them into the ground truth phonemes. In the more difficult second track, we provide the ground truth number of phonemes, and the evaluation assumes a \textit{one-to-one} mapping between units and phonemes. Evaluated systems may work from scratch or start from pretrained models. Pretrained models may be trained with or without supervision. However, no model may have initially seen any of the benchmark languages provided for the challenge, either \dev or \test ---including English. Submissions may only use the \dev languages to develop their architectures and fix hyperparameters.

We present the datasets, evaluation metrics, and results on baselines, which we open-source for participants to improve upon. The benchmark is available at \ifcameraready\url{https://benchmarks.cognitive-ml.fr/discophon}\else an anonymous URL\fi. 
 
\section{Datasets}
\benchmark includes 6 \dev languages (German, Swahili, Tamil, Thai, Turkish, Ukrainian) and 6 \test languages (Basque, English, French, Japanese, Mandarin Chinese, Wolof), selected to span a diverse range of phonemic categories, as shown in \Cref{tab:phonemes}, while ensuring sufficient data availability. All data consists of read speech, either from audiobooks or from sentence-level prompts. For each language, we provide \SI{10}{\hour} of training data, along with \SI{2}{\hour} \dev and \SI{2}{\hour} \test splits annotated with phone-level automatic transcriptions. Additional \SI{10}{\minute} and \SI{1}{\hour} train splits are included to study learning under more constrained conditions, following settings from previous benchmarks \cite{shi23g_interspeech}, without textual supervision. Speakers do not overlap across splits, and each evaluation set is gender-balanced with 10 male and 10 female speakers, in quantities as uniform as possible.

German, English, French, and Wolof data come from the ZeroSpeech 2017 challenge \cite{dunbar2017zeroa}, with Wolof extended using the original dataset \cite{gauthier2016collectinga}. German, English, and French are based on LibriVox\footnote{\url{https://librivox.org}} audiobooks, from which we selected 10 male and 10 female speakers per train set. We use the original Kaldi-based alignments \cite{povey2011kaldi} for all four languages. The remaining languages are sourced from Common Voice \cite{ardila2020commona}, which involves a much larger number of speakers per language since contributors read individual sentences rather than full book chapters. For these, we use alignments from the VoxCommunis project \cite{ahn2022voxcommunis}, produced with the Montreal Forced Aligner \cite{mcauliffe2017montreal}, and simplify some specific phonetic notations by folding them back to a single IPA symbol representing the underlying contrastive category.

We distribute annotations and splits for all languages, along with audio data for the ZeroSpeech languages. Audio for CommonVoice languages must be downloaded separately.

\begin{table*}[ht]
    \addtolength{\tabcolsep}{-0.25em}
    \centering
    \caption{\textbf{\benchmark baselines many-to-one scores (256 units).} Units are mapped to the most probable phoneme for evaluation. Best layer L is chosen to minimize continuous ABX on \dev languages. HuBERT units: $K$-means on layer L trained on pretraining data (zero-shot), or finetuning data; SpidR units: prediction heads of layer L. Results (in $\%$) averaged across \dev and \test languages, resp. ABX on continuous representations averaged between within and across speaker conditions. Best scores in \textbf{bold}.\label{tab:bench-many-to-one}}
    \begin{tabular}{lcccccccccc}
\toprule
& \multicolumn{5}{c}{\dev languages} & \multicolumn{5}{c}{\test languages} \\
\cmidrule(lr){2-6} \cmidrule(lr){7-11}
& \bf PER $\downarrow$  & \bf$\bm R$-value $\uparrow$  & $F_1$ $\uparrow$  & PNMI $\uparrow$  & ABX c. $\downarrow$  & \bf PER $\downarrow$  & \bf$\bm R$-value $\uparrow$  & $F_1$ $\uparrow$  & PNMI $\uparrow$  & ABX c. $\downarrow$ \\
\midrule
\textbf{Zero-shot} \\
HuBERT MMS-ulab (L12) & 117.25 & 17.70 & 58.67 & 48.32 & 11.18 & 126.46 & 6.99 & 57.79 & 51.05 & 8.72 \\
HuBERT VP-20 (L11) & 118.26 & 19.63 & 58.87 & 47.01 & 10.23 & 127.02 & 9.39 & 58.18 & 49.49 & 7.79 \\
SpidR MMS-ulab (L5) & 80.50 & 45.93 & $\mathbf{64.77}$ & $\mathbf{56.23}$ & 9.30 & 84.81 & 40.99 & $\mathbf{65.65}$ & $\mathbf{58.58}$ & 7.37 \\
SpidR VP-20 (L6) & $\mathbf{64.41}$ & $\mathbf{58.75}$ & 61.83 & 54.51 & $\mathbf{8.80}$ & $\mathbf{64.48}$ & $\mathbf{56.04}$ & 63.05 & 58.12 & $\mathbf{6.57}$ \\
\addlinespace
\textbf{Finetuned on 10h} \\
HuBERT MMS-ulab (L12) & 111.90 & 18.99 & 58.59 & 51.19 & 9.26 & 113.55 & 15.01 & 59.68 & 54.43 & 6.93 \\
HuBERT VP-20 (L11) & 95.90 & 30.48 & 61.45 & 54.66 & 7.68 & 98.02 & 24.37 & 61.16 & 57.56 & 5.49 \\
SpidR MMS-ulab (L5) & 65.92 & 54.38 & $\mathbf{67.99}$ & $\mathbf{61.90}$ & 8.65 & 65.44 & 51.57 & $\mathbf{68.90}$ & $\mathbf{64.61}$ & 6.40 \\
SpidR VP-20 (L6) & $\mathbf{63.06}$ & $\mathbf{55.32}$ & 63.79 & 61.20 & $\mathbf{6.62}$ & $\mathbf{61.67}$ & $\mathbf{53.27}$ & 64.35 & 64.38 & $\mathbf{5.09}$ \\
\bottomrule
\end{tabular}
\end{table*}

\begin{table}[ht]
    \addtolength{\tabcolsep}{-0.47em}
    \centering
    \caption{\textbf{\benchmark baselines one-to-one scores ($|\mathcal{P}|+1$ units).} Each phoneme is mapped to a single unit. For each model and language, $K$-means is fitted on pretraining (zero-shot models) or finetuning data. Target layer as in \Cref{tab:bench-many-to-one}, best scores in \textbf{bold}.\label{tab:bench-one-to-one}}
    \begin{tabular}{lcccc}
\toprule
& \multicolumn{2}{c}{\dev languages} & \multicolumn{2}{c}{\test languages} \\
\cmidrule(lr){2-3} \cmidrule(lr){4-5}
& \bf PER $\downarrow$  & \bf$\bm R$-value $\uparrow$  & \bf PER $\downarrow$  & \bf$\bm R$-value $\uparrow$ \\
\midrule
\textbf{Zero-shot} \\
HuBERT MMS-ulab & 271.80 & -99.10 & 264.84 & -93.39 \\
HuBERT VP-20 & 184.11 & -27.71 & 196.58 & -37.36 \\
SpidR MMS-ulab & 135.64 & 14.72 & 142.06 & 7.13 \\
SpidR VP-20 & $\mathbf{113.57}$ & $\mathbf{31.53}$ & $\mathbf{120.03}$ & $\mathbf{24.45}$ \\
\addlinespace
\textbf{Finetuned on 10h} \\
HuBERT MMS-ulab & 204.44 & -43.25 & 192.01 & -35.34 \\
HuBERT VP-20 & 174.33 & -25.25 & 178.67 & -28.67 \\
SpidR MMS-ulab & $\mathbf{128.53}$ & $\mathbf{16.28}$ & $\mathbf{133.47}$ & $\mathbf{8.22}$ \\
SpidR VP-20 & 136.11 & 8.67 & 136.62 & 3.74 \\
\bottomrule
\end{tabular}
\end{table}

\section{Evaluation}
We define two tracks based on the vocabulary size of the model. In the \textit{many-to-one} track, it exceeds the number of phonemes: multiple units may map to the same category. We ask benchmark participants to submit systems with 256 units, for fair comparison. In the \textit{one-to-one} track, the vocabulary size equals the number of phonemes plus one (for silence).

We first assess unit quality with PNMI \cite{hsu2021hubert}. We then derive a mapping from units to phonemes using the gold annotations, as illustrated in \Cref{fig:streams}. This produces a phonetic transcription, which we evaluate for recognition with \textbf{Phone Error Rate (PER)} and segmentation with $\bm R$\textbf{-value} and $F_1$.

\subsection{Mapping units to phonemes}

Let $\bm{u} = (u_1, \dots, u_T)$ be the sequence of discrete units to evaluate, corresponding to a full dataset split, and let $\bm{p} = (p_1, \dots, p_T)$ be the corresponding sequence of gold phones, where $T$ denotes the number of time steps at the resolution of the evaluated system. We denote by $\mathcal{P}$ the predefined set of phonemes, and by $\mathcal{U}$ the one of units. The empirical joint distribution of phones and units is
\begin{equation}
 \prob(i, j) = \frac{1}{T} \sum_{t=1}^T \left(p_t = i \wedge u_t = j\right), i \in \mathcal{P}, j \in \mathcal{U}.
\end{equation}

The many-to-one assignment maps each unit to its most frequent phoneme $A: j \mapsto \arg \max_{i \in \mathcal{P}} \prob(i, j)$.
The assigned sequence $\bm{a} = (a_1, \dots, a_T)$ is obtained by applying this mapping at each time step: $a_t = A(u_t)$. For the one-to-one assignment, we impose each phoneme to be mapped to a single unit: $A$ has to be a bijection. We derive it by solving the linear assignment problem that maximizes $\prob(i, j)$ with SciPy.

Setting the vocabulary size is crucial for fair comparison: with this setup, an unconstrained many-to-one mapping can be improved by increasing $|\mathcal{U}|$. In the extreme case where $|\mathcal{U}| = T$ and where each unit appears exactly once, the mapping would be perfect. A fixed vocabulary size eliminates this confound.

\subsection{Evaluation metrics}
\paragraph{Unit quality.} The PNMI between $\bm{p}$ and $\bm{u}$ is:
\begin{equation}
    \text{PNMI}(\bm{p}, \bm{u}) = \frac{I(\bm{p};\bm{u})}{H(\bm{p})} = \frac{\sum_{i,j}\prob(i, j)\log \frac{\prob(i, j)}{\prob_{\bm{p}}(i) \prob_{\bm{u}}(j)}}{- \sum_i \prob_{\bm{p}}(i) \log \prob_{\bm{p}}(i)},
\end{equation}

where $\prob_{\bm{p}}(i) = \sum_{j \in \mathcal{U}} \prob(i, j)$ and $\prob_{\bm{u}}(j) = \sum_{i \in \mathcal{P}} \prob(i, j)$ are the  marginal distributions. It measures the fraction of phone entropy explained by the discrete units.

\paragraph{Recognition.} PNMI is sensitive both to unit quality and alignment. Since the mapping produces a phone sequence $\bm{a}$ from the units $\bm{u}$, we compare it directly to the gold sequence $\bm{p}$ by computing the \textbf{PER} to abstract away from the alignment.

\paragraph{Segmentation.} Recognition evaluates predicted labels but not their temporal alignment. Segmentation evaluation is complementary: it ignores labels and only compares boundary positions.

We report $F_1$ and $\bm R$\textbf{-value} \cite{rasanen2009improveda}, which penalizes over-segmentation more than $F_1$. Following \cite{rasanen2009improveda}, we allow a tolerance of $\pm$20 ms around each gold boundary and split overlapping windows at their midpoint.

\paragraph{Discriminability.} We provide utilities to optionally compute ABX discriminability \cite{schatz13_interspeech,schatz:tel-01407461} on continuous representations of triphones or discrete units, using the fastabx library \cite{fastabx}. ABX measures whether two instances of the same triphone (e.g., \textipa{/bag/}) are closer to one another in embedding space than to instances of a minimally contrasting triphone (e.g., \textipa{/beg/}). On discrete units, ABX is related to PNMI but with a hard threshold for success: two realizations of the same triphone must be encoded as the same sequence. On continuous representations, it is a useful proxy during development to guide pretraining and select intermediate layers with easily available phonetic information without requiring discretization.

\section{Baselines}
\begin{figure*}
    \centering \includegraphics{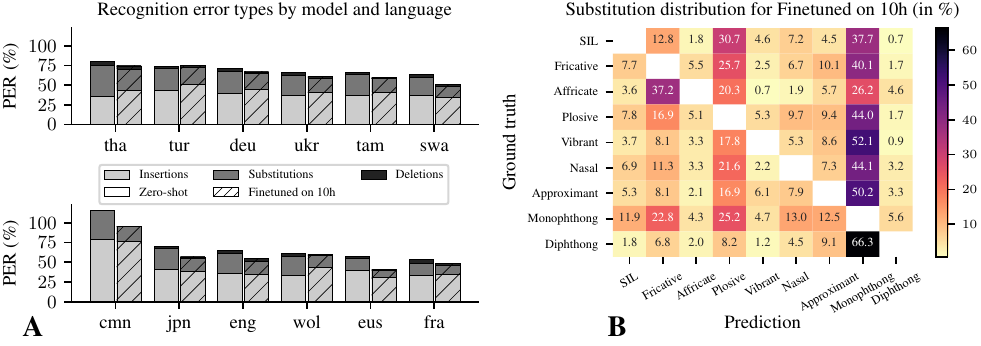}
    \caption{\textbf{Analysis of recognition errors in a many-to-one evaluation.} {\bf A}: Breakdown of the phone error rate of SpidR VP-20 with 256 units into insertions, substitutions, and deletions for each language, comparing zero-shot and finetuned models. {\bf B}: Substitution confusion matrix between phoneme class for the finetuned model, averaged across languages. Each row indicates, given a ground-truth class of phonemes, which predicted class it was replaced by (in $\%$). \label{fig:details}}
    \vspace{-0.7em}
\end{figure*}

\subsection{Models}
Since the benchmark evaluates phoneme discovery in unseen languages, systems must not have been exposed to any \dev or \test languages during pretraining. This notably excludes all models pretrained on English. We therefore provide four baselines pretrained on multilingual datasets that satisfy this constraint, using two architectures: HuBERT \cite{hsu2021hubert} and SpidR \cite{poli2025spidr}. HuBERT has been extensively shown to encode phonetic information in its intermediate representations \cite{wells2022phonetic}. SpidR was found to make it more readily accessible, leading to improvements in downstream spoken language modeling \cite{poli2025spidr}.

\subsection{Pretraining and finetuning}
We pretrain each model on either VP-20 or MMS-ulab-v2. VP-20 is a custom subset of VoxPopuli \cite{wang2021voxpopuli} consisting of 6k hours of speech sampled uniformly across all languages except English, French, and German. MMS-ulab-v2 \cite{chen2024robusta,pratap2024scaling} is a massively multilingual corpus spanning 4,023 ISO3 languages sourced from the Global Recordings Network. We apply voice activity detection \cite{bredin23_interspeech}, segment utterances to a maximum of \SI{30}{\second}, and remove benchmark languages as well as closely related ones (e.g., Lao due to Thai, all Chinese languages, etc.). The resulting dataset contains 8k hours in 3,966 languages.

These two datasets have comparable size but opposite distributions: VP-20 provides substantial data per language in equal proportions, while MMS-ulab-v2 has broad coverage but half of its languages contain less than \SI{10}{min} of audio. 

SpidR models are pretrained with the original hyperparameters. For HuBERT, we extract MFCCs, train $K$-means with $K=100$ on 1\% of the pretraining data, and run the first iteration with default hyperparameters on 16 GPUs. We select the best layer by ABX on the \dev languages (VP-20: L10; MMS-ulab: L9), then train the second iteration with $K=500$ and again select the best layer (VP-20: L11; MMS-ulab: L12).

We finetune on each language independently using the same configuration. Finetuning consists in continuing pretraining on the target language data with a new learning rate schedule. For SpidR, discrete units are derived from the prediction heads. For HuBERT, we train $K$-means on the best layer on the finetuning data, then, after finetuning, we select the best layer again and fit a final $K$-means to get the evaluation units. For zero-shot evaluation, the $K$-means comes from the pretraining set.

\subsection{Results}
\paragraph{Many-to-one.} We report many-to-one results for our four baselines with 256 units in \Cref{tab:bench-many-to-one}, in both zero-shot and finetuned conditions. Units are derived either via $K$-means (HuBERT) or prediction heads (SpidR). SpidR consistently outperforms HuBERT across all metrics and conditions. The gap in segmentation between HuBERT and SpidR is driven more by $R$-value than by $F_1$, indicating that HuBERT is more prone to over-segmentation, producing unit spans with spurious insertions that do not align with a single phoneme.

\paragraph{One-to-one.} The one-to-one evaluation is substantially harder, as shown in \Cref{tab:bench-one-to-one}. The difficulty does not stem from the reduced number of units, but from the strict assignment constraint: every unit maps to a distinct phoneme. In the many-to-one setting, frequent targets like \textipa{/a/}, \textipa{/i/}, or \sil naturally co-occur with multiple units, while rare phonemes are covered by only a few. With a one-to-one mapping, however, some units are inevitably assigned to phonemes they poorly represent, degrading both recognition and segmentation. A possible direction would be to hierarchically group units from a larger vocabulary down to the target size.

\paragraph{Recognition errors.} As shown in \Cref{fig:details}.\textbf{A}, PER varies widely across languages, from $41.34\%$ in Basque to $95.98\%$ in Mandarin Chinese for the finetuned model. Insertions dominate the error distribution (zero-shot: $58\%$, finetuned: $69\%$), followed by substitutions ($37\%$, resp. $28\%$) and deletions ($5\%$, resp. $4\%$). Finetuning primarily reduces substitutions but has little effect on insertions. This suggests that methods that smooth predictions across frames during unit derivation could yield substantial gains by limiting spurious transitions. \Cref{fig:details}.\textbf{B} shows, for each phoneme class, which class the predicted phoneme belongs to when a substitution error occurs, aggregated across all languages. A large part of substitution errors result in monophthongs being incorrectly predicted, even for ground truth plosives or fricatives.

\section{Conclusion}
We introduced \benchmark, a multilingual benchmark for evaluating phoneme discovery from discrete speech units. The benchmark defines two tracks, \textit{many-to-one} (256 units) and \textit{one-to-one} (number of phonemes plus one), and evaluates unit quality, recognition, and segmentation. Since systems may not be pretrained on any benchmark language, we provide four pretrained SpidR and HuBERT baselines to facilitate experimentation. While much attention has been paid to whether unsupervised speech models’ (continuous or discrete) representations are predictive of phonemes, this benchmark, for the first time, explicitly fixes the goal of learning a discrete set of categories that are in one-to-one correspondence with the phoneme inventory of a language.

We see two main directions for future work on the benchmark itself. The first is to broaden linguistic diversity, in particular by including languages with non-pulmonic consonants, clicks, and other underrepresented contrasts. The second is to enrich the levels of description. Our current annotations are pseudo-phonemic, derived from automatic transcriptions that reflect canonical pronunciations and may omit crucial phonological distinctions, such as tones in Mandarin Chinese. Finer-grained phonetic annotations would help clarify whether some aspects of the learned representations are more phonetic or phonemic in nature. Looking ahead, the evaluation can be extended beyond phonemes to larger units such as syllables, and even words.

\clearpage
\ifcameraready
\section{Acknowledgements}
This work was performed using HPC resources from GENCI-IDRIS (Grant 2023-AD011014368) and was supported in part by the Agence Nationale pour la Recherche (ANR-17-EURE-0017 Frontcog, ANR10-IDEX-0001-02 PSL*). MP acknowledges Ph.D. funding from Agence de l’Innovation de Défense. ED and MK were funded by an ERC grant (InfantSimulator, 101142705). Views and opinions expressed are those of the authors only and do not necessarily reflect those of the European Union or the European Research Council. Neither the European Union nor the granting authority can be held responsible for them.
\fi

\section{Generative AI Use Disclosure}
Generative AI was used to polish the manuscript.

\bibliographystyle{IEEEtran}
\bibliography{mybib}

@phdthesis{schatz:tel-01407461,
	title = {{ABX-Discriminability} Measures and Applications},
	author = {Schatz, Thomas},
	url = {https://hal.science/tel-01407461},
	school = {Universit{\'e} Paris 6 (UPMC)},
	year = {2016},
	month = {Sep},
	type = {Theses},
	hal_id = {tel-01407461},
	hal_version = {v1},
}

@article{poli2025spidr,
	title = {{{SpidR}}: {{Learning Fast}} and {{Stable Linguistic Units}} for {{
	         Spoken Language Models Without Supervision}}},
	shorttitle = {{{SpidR}}},
	author = {Poli, Maxime and Luthra, Mahi and Benchekroun, Youssef and Higuchi
	          , Yosuke and Gleize, Martin and Shen, Jiayi and Algayres, Robin and
	          Chung, Yu-An and Assran, Mido and Pino, Juan and Dupoux, Emmanuel},
	year = 2025,
	month = jul,
	journal = {Transactions on Machine Learning Research},
	issn = {2835-8856},
	url = {https://openreview.net/forum?id=E7XAFBpfZs},
	langid = {english},
}

@inproceedings{shi23g_interspeech,
	title = {{ML-SUPERB: Multilingual Speech Universal PERformance Benchmark}},
	author = {Jiatong Shi and Dan Berrebbi and William Chen and En-Pei Hu and
	          Wei-Ping Huang and Ho-Lam Chung and Xuankai Chang and Shang-Wen Li
	          and Abdelrahman Mohamed and Hung-yi Lee and Shinji Watanabe},
	year = {2023},
	booktitle = {{Interspeech 2023}},
	pages = {884--888},
	doi = {10.21437/Interspeech.2023-1316},
	issn = {2958-1796},
}

@inproceedings{schatz13_interspeech,
	title = {Evaluating speech features with the minimal-pair {ABX} task: {
	         Analysis} of the classical {MFC/PLP} pipeline},
	author = {Thomas Schatz and Vijayaditya Peddinti and Francis Bach and Aren
	          Jansen and Hynek Hermansky and Emmanuel Dupoux},
	year = {2013},
	booktitle = {Interspeech 2013},
	pages = {1781--1785},
	doi = {10.21437/Interspeech.2013-441},
	issn = {2958-1796},
}

@misc{fastabx,
	title = {fastabx: {A} library for efficient computation of {ABX}
	         discriminability},
	author = {Maxime Poli and Emmanuel Chemla and Emmanuel Dupoux},
	year = {2025},
	eprint = {2505.02692},
	archiveprefix = {arXiv},
	primaryclass = {cs.CL},
	url = {https://arxiv.org/abs/2505.02692},
}

@inproceedings{chen2024robusta,
	title = {Towards {{Robust Speech Representation Learning}} for {{Thousands}}
	         of {{Languages}}},
	booktitle = {Proceedings of the 2024 {{Conference}} on {{Empirical Methods}}
	             in {{Natural Language Processing}}},
	author = {Chen, William and Zhang, Wangyou and Peng, Yifan and Li, Xinjian
	          and Tian, Jinchuan and Shi, Jiatong and Chang, Xuankai and Maiti,
	          Soumi and Livescu, Karen and Watanabe, Shinji},
	editor = {{Al-Onaizan}, Yaser and Bansal, Mohit and Chen, Yun-Nung},
	year = 2024,
	month = nov,
	pages = {10205--10224},
	publisher = {Association for Computational Linguistics},
	doi = {10.18653/v1/2024.emnlp-main.570},
}

@inproceedings{poli2024improving,
	title = {Improving {{Spoken Language Modeling}} with {{Phoneme
	         Classification}}: {{A Simple Fine-tuning Approach}}},
	shorttitle = {Improving {{Spoken Language Modeling}} with {{Phoneme
	              Classification}}},
	booktitle = {Proceedings of the 2024 {{Conference}} on {{Empirical Methods}}
	             in {{Natural Language Processing}}},
	author = {Poli, Maxime and Chemla, Emmanuel and Dupoux, Emmanuel},
	year = 2024,
	month = nov,
	pages = {5284--5292},
	publisher = {Association for Computational Linguistics},
	doi = {10.18653/v1/2024.emnlp-main.302},
	copyright = {All rights reserved},
}

@inproceedings{yeh2024estimating,
  title = {Estimating the Completeness of Discrete Speech Units},
  booktitle = {2024 {{IEEE Spoken Language Technology Workshop}} ({{SLT}})},
  author = {Yeh, Sung-Lin and Tang, Hao},
  year = 2024,
  pages = {415--422},
  publisher = {IEEE},
  url = {https://ieeexplore.ieee.org/abstract/document/10832198/}
}

@article{pratap2024scaling,
  title={Scaling speech technology to 1,000+ languages},
  author={Pratap, Vineel and Tjandra, Andros and Shi, Bowen and Tomasello, Paden and Babu, Arun and Kundu, Sayani and Elkahky, Ali and Ni, Zhaoheng and Vyas, Apoorv and Fazel-Zarandi, Maryam and others},
  journal={Journal of Machine Learning Research},
  volume={25},
  number={97},
  pages={1--52},
  year={2024}
}

@inproceedings{wang2021voxpopuli,
	title = {{{VoxPopuli}}: {{A Large-Scale Multilingual Speech Corpus}} for {{
	         Representation Learning}}, {{Semi-Supervised Learning}} and {{
	         Interpretation}}},
	shorttitle = {{{VoxPopuli}}},
	booktitle = {Proceedings of the 59th {{Annual Meeting}} of the {{Association
	             }} for {{Computational Linguistics}} and the 11th {{
	             International Joint Conference}} on {{Natural Language
	             Processing}} ({{Volume}} 1: {{Long Papers}})},
	author = {Wang, Changhan and Riviere, Morgane and Lee, Ann and Wu, Anne and
	          Talnikar, Chaitanya and Haziza, Daniel and Williamson, Mary and
	          Pino, Juan and Dupoux, Emmanuel},
	year = 2021,
	month = aug,
	pages = {993--1003},
	publisher = {Association for Computational Linguistics},
	doi = {10.18653/v1/2021.acl-long.80},
}

@article{hsu2021hubert,
	title = {{{HuBERT}}: {{{Self-Supervised} Speech Representation Learning}} by
	         {{Masked Prediction}} of {{Hidden Units}}},
	shorttitle = {HuBERT},
	author = {Hsu, Wei-Ning and Bolte, Benjamin and Tsai, Yao-Hung Hubert and
	          Lakhotia, Kushal and Salakhutdinov, Ruslan and Mohamed, Abdelrahman
	          },
	year = {2021},
	journal = {IEEE/ACM Transactions on Audio, Speech, and Language Processing},
	volume = {29},
	pages = {3451--3460},
	issn = {2329-9304},
	doi = {10.1109/TASLP.2021.3122291},
}

@article{dunbar2022selfsupervised,
	title = {Self-{{Supervised Language Learning From Raw Audio}}: {{{Lessons}
	         From}} the {{Zero Resource Speech Challenge}}},
	shorttitle = {Self-{{Supervised Language Learning From Raw Audio}}},
	author = {Dunbar, Ewan and Hamilakis, Nicolas and Dupoux, Emmanuel},
	year = {2022},
	month = {oct},
	journal = {IEEE Journal of Selected Topics in Signal Processing},
	volume = {16},
	number = {6},
	pages = {1211--1226},
	issn = {1941-0484},
	doi = {10.1109/JSTSP.2022.3206084},
}

@inproceedings{dunbar2021zero,
	title = {The {{Zero Resource Speech Challenge}} 2021: {{{Spoken} Language
	         Modelling}}},
	shorttitle = {The {{Zero Resource Speech Challenge}} 2021},
	booktitle = {Interspeech 2021},
	author = {Dunbar, Ewan and Bernard, Mathieu and Hamilakis, Nicolas and
	          Nguyen, Tu Anh and Seyssel, Maureen De and Roz{\'e}, Patricia and
	          Rivi{\`e}re, Morgane and Kharitonov, Eugene and Dupoux, Emmanuel},
	year = {2021},
	month = {aug},
	pages = {1574--1578},
	doi = {10.21437/Interspeech.2021-1755},
	langid = {english},
}

@misc{nguyen2020zeroresourcespeechbenchmark,
	title = {The Zero Resource Speech Benchmark 2021: {Metrics} and baselines
	         for unsupervised spoken language modeling},
	author = {Tu Anh Nguyen and Maureen de Seyssel and Patricia Rozé and Morgane
	          Rivière and Evgeny Kharitonov and Alexei Baevski and Ewan Dunbar
	          and Emmanuel Dupoux},
	year = {2020},
	eprint = {2011.11588},
	archiveprefix = {arXiv},
	primaryclass = {cs.CL},
	url = {https://arxiv.org/abs/2011.11588},
}

@inproceedings{choi24b_interspeech,
	title = {Self-Supervised Speech Representations are More Phonetic than
	         Semantic},
	author = {Kwanghee Choi and Ankita Pasad and Tomohiko Nakamura and Satoru
	          Fukayama and Karen Livescu and Shinji Watanabe},
	year = {2024},
	booktitle = {Interspeech 2024},
	pages = {4578--4582},
	doi = {10.21437/Interspeech.2024-1157},
	issn = {2958-1796},
}

@inproceedings{deseyssel22_interspeech,
	title = {Probing phoneme, language and speaker information in unsupervised
	         speech representations},
	author = {Maureen {de Seyssel} and Marvin Lavechin and Yossi Adi and
	          Emmanuel Dupoux and Guillaume Wisniewski},
	year = {2022},
	booktitle = {Interspeech 2022},
	pages = {1402--1406},
	doi = {10.21437/Interspeech.2022-373},
	issn = {2958-1796},
}

@article{park2008unsupervised,
	title = {Unsupervised {{Pattern Discovery}} in {{Speech}}},
	author = {Park, Alex and Glass, James},
	year = 2008,
	month = feb,
	journal = {Audio, Speech, and Language Processing, IEEE Transactions on},
	volume = {16},
	pages = {186--197},
	doi = {10.1109/TASL.2007.909282},
}

@article{clements1985geometry,
	title = {The Geometry of Phonological Features},
	author = {Clements, G. N.},
	year = 1985,
	month = may,
	journal = {Phonology},
	volume = {2},
	number = {1},
	pages = {225--252},
	issn = {2059-6286, 0265-8062},
	doi = {10.1017/S0952675700000440},
	langid = {english},
}

@article{kempton2014discovering,
	title = {Discovering the Phoneme Inventory of an Unwritten Language: {{A}}
	         Machine-Assisted Approach},
	shorttitle = {Discovering the Phoneme Inventory of an Unwritten Language},
	author = {Kempton, Timothy and Moore, Roger K.},
	year = 2014,
	month = jan,
	journal = {Speech Communication},
	volume = {56},
	pages = {152--166},
	issn = {01676393},
	doi = {10.1016/j.specom.2013.02.006},
	langid = {english},
}

@book{moseley2010atlas,
	title = {Atlas of the World's Languages in Danger},
	author = {Moseley, Christopher},
	year = {2010},
	publisher = {Unesco},
}

@inproceedings{bredin23_interspeech,
	title = {{pyannote.audio 2.1 speaker diarization pipeline: principle,
	         benchmark, and recipe}},
	author = {Hervé Bredin},
	year = {2023},
	booktitle = {{Interspeech 2023}},
	pages = {1983--1987},
	doi = {10.21437/Interspeech.2023-105},
	issn = {2958-1796},
}

@inproceedings{rasanen2009improveda,
	title = {An Improved Speech Segmentation Quality Measure: The r-Value},
	shorttitle = {An Improved Speech Segmentation Quality Measure},
	booktitle = {Interspeech 2009},
	author = {R{\"a}s{\"a}nen, Okko Johannes and Laine, Unto Kalervo and
	          Altosaar, Toomas},
	year = 2009,
	month = sep,
	pages = {1851--1854},
	publisher = {ISCA},
	doi = {10.21437/Interspeech.2009-538},
	langid = {english},
}

@inproceedings{ahn2022voxcommunis,
	title = {{{VoxCommunis}}: {{A Corpus}} for {{Cross-linguistic Phonetic
	         Analysis}}},
	shorttitle = {{{VoxCommunis}}},
	booktitle = {Proceedings of the {{Thirteenth Language Resources}} and {{
	             Evaluation Conference}}},
	author = {Ahn, Emily and Chodroff, Eleanor},
	year = 2022,
	month = jun,
	pages = {5286--5294},
	publisher = {European Language Resources Association},
	url = {https://aclanthology.org/2022.lrec-1.566/},
}

@inproceedings{ardila2020commona,
	title = {Common {{Voice}}: {{A Massively-Multilingual Speech Corpus}}},
	shorttitle = {Common {{Voice}}},
	booktitle = {Proceedings of the {{Twelfth Language Resources}} and {{
	             Evaluation Conference}}},
	author = {Ardila, Rosana and Branson, Megan and Davis, Kelly and Kohler,
	          Michael and Meyer, Josh and Henretty, Michael and Morais, Reuben
	          and Saunders, Lindsay and Tyers, Francis and Weber, Gregor},
	year = 2020,
	month = may,
	pages = {4218--4222},
	publisher = {European Language Resources Association},
	url = {https://aclanthology.org/2020.lrec-1.520/},
	isbn = {979-10-95546-34-4},
	langid = {english},
}

@inproceedings{ludusan-etal-2014-bridging,
	title = "Bridging the gap between speech technology and natural language
	         processing: an evaluation toolbox for term discovery systems",
	author = "Ludusan, Bogdan and Versteegh, Maarten and Jansen, Aren and
	          Gravier, Guillaume and Cao, Xuan-Nga and Johnson, Mark and Dupoux,
	          Emmanuel",
	booktitle = "Proceedings of the Ninth International Conference on Language
	             Resources and Evaluation ({LREC}'14)",
	month = may,
	year = "2014",
	publisher = "European Language Resources Association (ELRA)",
	url = "https://aclanthology.org/L14-1284/",
	pages = "560--567",
}

@inproceedings{dunbar2017zeroa,
	title = {The Zero Resource Speech Challenge 2017},
	booktitle = {2017 {{IEEE Automatic Speech Recognition}} and {{Understanding
	             Workshop}} ({{ASRU}})},
	author = {Dunbar, Ewan and Cao, Xuan Nga and Benjumea, Juan and Karadayi,
	          Julien and Bernard, Mathieu and Besacier, Laurent and Anguera,
	          Xavier and Dupoux, Emmanuel},
	year = 2017,
	month = dec,
	pages = {323--330},
	doi = {10.1109/ASRU.2017.8268953},
}

@inproceedings{gauthier2016collectinga,
	title = {Collecting {{Resources}} in {{Sub-Saharan African Languages}} for {
	         {Automatic Speech Recognition}}: A {{Case Study}} of {{Wolof}}},
	shorttitle = {Collecting {{Resources}} in {{Sub-Saharan African Languages}}
	              for {{Automatic Speech Recognition}}},
	booktitle = {Proceedings of the {{Tenth International Conference}} on {{
	             Language Resources}} and {{Evaluation}} ({{LREC}}'16)},
	author = {Gauthier, Elodie and Besacier, Laurent and Voisin, Sylvie and
	          Melese, Michael and Elingui, Uriel Pascal},
	year = 2016,
	month = may,
	pages = {3863--3867},
	publisher = {European Language Resources Association (ELRA)},
	url = {https://aclanthology.org/L16-1611/},
}

@inproceedings{mcauliffe2017montreal,
	title = {Montreal {{Forced Aligner}}: {{Trainable Text-Speech Alignment
	         Using Kaldi}}},
	shorttitle = {Montreal {{Forced Aligner}}},
	booktitle = {Proc. {{Interspeech}} 2017},
	author = {McAuliffe, Michael and Socolof, Michaela and Mihuc, Sarah and
	          Wagner, Michael and Sonderegger, Morgan},
	year = 2017,
	pages = {498--502},
	doi = {10.21437/Interspeech.2017-1386},
}

@inproceedings{wells2022phonetic,
	title = {Phonetic {{Analysis}} of {{Self-supervised Representations}} of {{
	         English Speech}}},
	booktitle = {Proc. {{Interspeech}} 2022},
	author = {Wells, Dan and Tang, Hao and Richmond, Korin},
	year = 2022,
	pages = {3583--3587},
	doi = {10.21437/Interspeech.2022-10884},
}

@inproceedings{wang-etal-2022-self,
	title = "Self-supervised Semantic-driven Phoneme Discovery for Zero-resource
	         Speech Recognition",
	author = "Wang, Liming and Feng, Siyuan and Hasegawa-Johnson, Mark and Yoo,
	          Chang",
	booktitle = "Proceedings of the 60th Annual Meeting of the Association for
	             Computational Linguistics (Volume 1: Long Papers)",
	month = may,
	year = "2022",
	publisher = "Association for Computational Linguistics",
	url = "https://aclanthology.org/2022.acl-long.553/",
	doi = "10.18653/v1/2022.acl-long.553",
	pages = "8027--8047",
}

@article{kamper2022word,
	title = {Word segmentation on discovered phone units with dynamic
	         programming and self-supervised scoring},
	author = {Kamper, Herman},
	journal = {IEEE/ACM Transactions on Audio, Speech, and Language Processing},
	volume = {31},
	pages = {684--694},
	year = {2022},
	publisher = {IEEE},
}

@article{kamper2016unsupervised,
	title = {Unsupervised word segmentation and lexicon discovery using acoustic
	         word embeddings},
	author = {Kamper, Herman and Jansen, Aren and Goldwater, Sharon},
	journal = {IEEE/ACM Transactions on Audio, Speech, and Language Processing},
	volume = {24},
	number = {4},
	pages = {669--679},
	year = {2016},
	publisher = {IEEE},
}

@inproceedings{7953148,
	author = {Müller, Markus and Franke, Jörg and Waibel, Alex and Stüker,
	          Sebastian},
	booktitle = {2017 IEEE International Conference on Acoustics, Speech and
	             Signal Processing (ICASSP)},
	title = {Towards phoneme inventory discovery for documentation of unwritten
	         languages},
	year = {2017},
	volume = {},
	number = {},
	pages = {5200-5204},
	doi = {10.1109/ICASSP.2017.7953148},
}

@inproceedings{povey2011kaldi,
	title = {The {{Kaldi Speech Recognition Toolkit}}},
	booktitle = {{{IEEE}} 2011 {{Workshop}} on {{Automatic Speech Recognition}}
	             and {{Understanding}}},
	author = {Povey, Daniel and Ghoshal, Arnab and Boulianne, Gilles and Burget,
	          Luka{\textasciicaron}s and Glembek, Ond{\textasciicaron}rej and
	          Goel, Nagendra and Hannemann, Mirko and Motl{\i}{\textasciicaron}
	          cek, Petr and Qian, Yanmin and Schwarz, Petr and Silovsky, Jan and
	          Stemmer, Georg and Vesely, Karel},
	year = 2011,
	month = dec,
	publisher = {IEEE Signal Processing Society},
	langid = {english},
}

@inproceedings{abdullah2023informationtheoretic,
	title = {An {{Information-Theoretic Analysis}} of {{Self-supervised Discrete
	         Representations}} of {{Speech}}},
	booktitle = {{{INTERSPEECH}} 2023},
	author = {Abdullah, Badr M. and Shaik, Mohammed Maqsood and M{\"o}bius,
	          Bernd and Klakow, Dietrich},
	year = 2023,
	month = aug,
	pages = {2883--2887},
	publisher = {ISCA},
	doi = {10.21437/Interspeech.2023-2131},
	langid = {english},
}

@article{lakhotia-etal-2021-generative,
    title = "On Generative Spoken Language Modeling from Raw Audio",
    author = "Lakhotia, Kushal  and
      Kharitonov, Eugene  and
      Hsu, Wei-Ning  and
      Adi, Yossi  and
      Polyak, Adam  and
      Bolte, Benjamin  and
      Nguyen, Tu-Anh  and
      Copet, Jade  and
      Baevski, Alexei  and
      Mohamed, Abdelrahman  and
      Dupoux, Emmanuel",
    journal = "Transactions of the Association for Computational Linguistics",
    volume = "9",
    year = "2021",
    publisher = "MIT Press",
    url = "https://aclanthology.org/2021.tacl-1.79/",
    doi = "10.1162/tacl_a_00430",
    pages = "1336--1354",
}

@article{nguyen-etal-2025-spirit,
    title = "{S}pi{R}it-{LM}: Interleaved Spoken and Written Language Model",
    author = "Nguyen, Tu Anh  and
      Muller, Benjamin  and
      Yu, Bokai  and
      Costa-jussa, Marta R.  and
      Elbayad, Maha  and
      Popuri, Sravya  and
      Ropers, Christophe  and
      Duquenne, Paul-Ambroise  and
      Algayres, Robin  and
      Mavlyutov, Ruslan  and
      Gat, Itai  and
      Williamson, Mary  and
      Synnaeve, Gabriel  and
      Pino, Juan  and
      Sagot, Beno{\^i}t  and
      Dupoux, Emmanuel",
    journal = "Transactions of the Association for Computational Linguistics",
    volume = "13",
    year = "2025",
    publisher = "MIT Press",
    url = "https://aclanthology.org/2025.tacl-1.2/",
    doi = "10.1162/tacl_a_00728",
    pages = "30--52",
}

@article{borsos2023audiolm,
  title={Audiolm: a language modeling approach to audio generation},
  author={Borsos, Zal{\'a}n and Marinier, Rapha{\"e}l and Vincent, Damien and Kharitonov, Eugene and Pietquin, Olivier and Sharifi, Matt and Roblek, Dominik and Teboul, Olivier and Grangier, David and Tagliasacchi, Marco and others},
  journal={IEEE/ACM transactions on audio, speech, and language processing},
  volume={31},
  pages={2523--2533},
  year={2023},
  publisher={IEEE}
}

@inproceedings{nguyen23_interspeech,
  title     = {{Expresso: A Benchmark and Analysis of Discrete Expressive Speech Resynthesis}},
  author    = {Tu Anh Nguyen and Wei-Ning Hsu and Antony D'Avirro and Bowen Shi and Itai Gat and Maryam Fazel-Zarani and Tal Remez and Jade Copet and Gabriel Synnaeve and Michael Hassid and Felix Kreuk and Yossi Adi and Emmanuel Dupoux},
  year      = {2023},
  booktitle = {{Interspeech 2023}},
  pages     = {4823--4827},
  doi       = {10.21437/Interspeech.2023-1905},
  issn      = {2958-1796},
}
\end{document}